%% file: main.tex
\documentclass{article}

\PassOptionsToPackage{numbers,compress}{natbib}

\usepackage[dblblindworkshop, final]{neurips_2026}

\input{math_commands.tex}

\usepackage[utf8]{inputenc} 
\usepackage[T1]{fontenc}    
\usepackage{hyperref}       
\usepackage{url}            
\usepackage{booktabs}       
\usepackage{amsfonts}       
\usepackage{amssymb}        
\usepackage{nicefrac}       
\usepackage{microtype}      
\usepackage[table]{xcolor} 
\usepackage{amsmath}        
\usepackage{subcaption}
\usepackage{graphicx}
\usepackage{tikz}
\usetikzlibrary{tikzmark}
\usepackage{makecell}
\usepackage{cleveref}
\usepackage{array}
\usepackage[most]{tcolorbox}

\usepackage{apptools}
\AtAppendix{
  \counterwithin{figure}{section}
  \counterwithin{table}{section}
}

\title{AutoIndexer: Flexible-Order Decoding by Training Causal Models on Chains of Edits}

\workshoptitle{BeNTo: Beyond Next‑Token Prediction -- Diffusion \& Flow Models for Next‑Generation Decoding}

\definecolor{apricot}{rgb}{0.98, 0.81, 0.69}

\newcommand{\cmark}{\checkmark}
\newcommand{\xmark}{$\times$}
\definecolor{normalbg}{HTML}{E5E7EB}
\definecolor{normalfg}{HTML}{374151}
\definecolor{editbg}{HTML}{D9F99D}
\definecolor{editfg}{HTML}{4D7C0F}
\definecolor{returnbg}{HTML}{FBCFE8}
\definecolor{returnfg}{HTML}{9D174D}
\definecolor{eosbg}{HTML}{FDE3C7}
\definecolor{eosfg}{HTML}{9A5B13}
\definecolor{corruptfg}{HTML}{B91C1C}
\definecolor{qualbg}{HTML}{F8FAFC}
\definecolor{qualframe}{HTML}{CBD5E1}
\definecolor{quallabelfg}{HTML}{64748B}
\definecolor{successbg}{HTML}{DCFCE7}
\definecolor{successfg}{HTML}{15803D}
\definecolor{partialbg}{HTML}{FEF3C7}
\definecolor{partialfg}{HTML}{B45309}
\definecolor{failurebg}{HTML}{FEE2E2}
\definecolor{failurefg}{HTML}{B91C1C}
\newcommand{\editmk}[1]{\colorbox{editbg}{\textcolor{editfg}{\ttfamily\bfseries[EDIT #1]}}}
\newcommand{\returnmk}{\colorbox{returnbg}{\textcolor{returnfg}{\ttfamily\bfseries[RETURN]}}}
\newcommand{\corr}[1]{\textcolor{corruptfg}{\texttt{<<<#1>>>}}}
\newtcolorbox{qualbox}{
  enhanced, breakable,
  colback=qualbg, colframe=qualframe,
  boxrule=0.4pt, arc=1.5mm,
  left=6pt, right=6pt, top=5pt, bottom=5pt,
  boxsep=0pt,
  before skip=4pt, after skip=4pt,
  fontupper=\small
}
\newcommand{\qualstatus}[3]{\colorbox{#1}{\textcolor{#2}{\textsf{\footnotesize\bfseries\enskip#3\enskip}}}}
\newcommand{\successtag}[1]{\qualstatus{successbg}{successfg}{#1}}
\newcommand{\partialtag}[1]{\qualstatus{partialbg}{partialfg}{#1}}
\newcommand{\failuretag}[1]{\qualstatus{failurebg}{failurefg}{#1}}
\newcommand{\quallev}[2]{\textsf{\footnotesize\textcolor{quallabelfg}{Levenshtein~\textbf{#1}$\,\to\,$\textbf{#2}}}}
\newcommand{\qualhead}[2]{\noindent#1\hfill#2\\[2pt]\noindent{\color{qualframe}\rule{\linewidth}{0.4pt}}\\[3pt]}
\newcommand{\quallabelwidth}{1.5cm}
\newcommand{\qualrow}[2]{\noindent\makebox[\quallabelwidth][l]{\textsf{\scriptsize\bfseries\textcolor{quallabelfg}{#1}}}\begin{minipage}[t]{\dimexpr\linewidth-\quallabelwidth\relax}\raggedright#2\end{minipage}\\[2pt]}

\newcommand{\normaltoken}{{\small\colorbox{normalbg}{\textcolor{normalfg}{\ttfamily\bfseries[NORMAL]}}}}
\newcommand{\edittoken}{{\small\colorbox{editbg}{\textcolor{editfg}{\ttfamily\bfseries[EDIT]}}}}
\newcommand{\returntoken}{{\small\colorbox{returnbg}{\textcolor{returnfg}{\ttfamily\bfseries[RETURN]}}}}
\newcommand{\eostoken}{{\small\colorbox{eosbg}{\textcolor{eosfg}{\ttfamily\bfseries[EOS]}}}}

\author{%
  \begin{tabular}[t]{@{}c@{\hspace{2.75em}}c@{\hspace{2.75em}}c@{}}
    \bfseries
    Shu Ishida & Aliasghar Khani & Tianyu Zhang \\[0.5ex]
    The Cong Luong & James Seale Smith & Adam Gaier \\[1em]
    \multicolumn{3}{c}{\mdseries Autodesk Research}
  \end{tabular}
}

\begin{document}

\maketitle

\vspace{-1em}
\begin{abstract}
Autoregressive decoding is less suited to non-causally ordered data, e.g., cross-references or future dependencies, since their outputs cannot be revised post hoc, unlike iterative human reasoning. Diffusion LLMs partially mitigate this limitation but still cannot flexibly insert or remove arbitrary numbers of tokens from a pre-generated context.
We design AutoIndexer as a generative model with a novel training and decoding mechanism that simulates a \textit{Chain of Edits}, allowing the model to flexibly modify its prior outputs in hindsight via token insertions, substitutions, and deletions. 
The model learns to predict marker tokens for opening and closing edits, simultaneously moving a cursor to positions in the existing token sequence, analogous to a text editor. 
Training sequences are obtained by perturbing ground-truth labels with inverse chained edit operations, and the model learns to edit by undoing those inverses.
The perturbation training allows AutoIndexer to be robust to erroneous context and adaptable to editing tasks such as code correction. 
We demonstrate that our method exhibits post-generation context-editing capabilities that most autoregressive and diffusion models lack without compromising performance on standard LLM benchmarks. The code is available at \url{https://github.com/AutodeskAILab/autoindexer}.
\end{abstract}

\begin{figure}[!h]
    \vspace{-1em}
     \centering
     \includegraphics[width=0.82\textwidth]{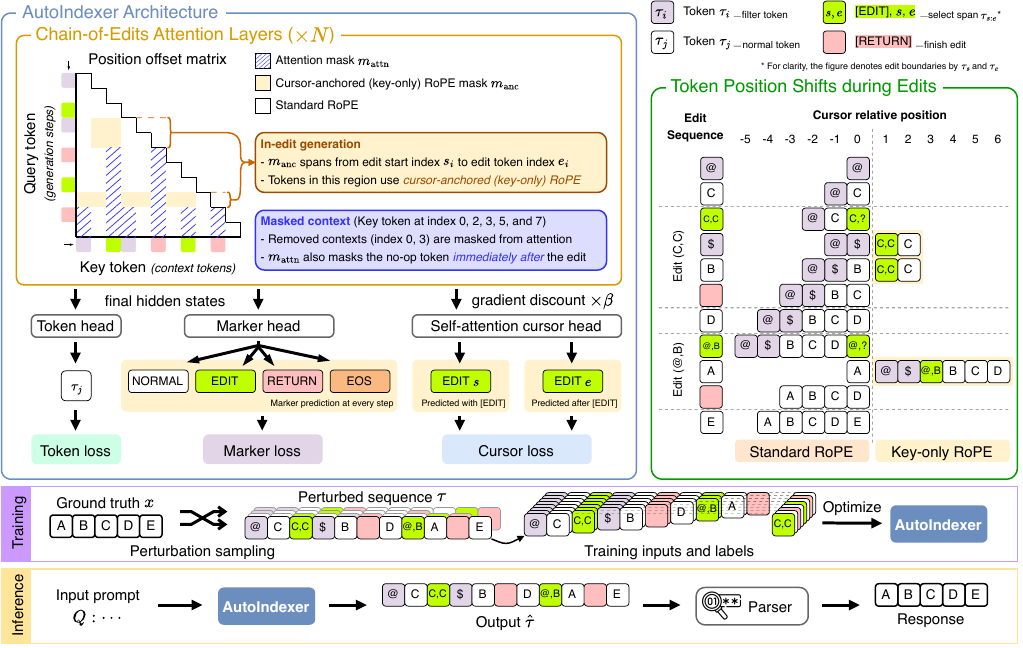}
    \caption{Overview of the AutoIndexer method.
    Analogous to using a text editor, AutoIndexer can \textit{edit} its prior outputs by predicting edit-start/-end cursor positions. The positional encodings of the tokens are modified after each edit and efficiently computed by our specialized attention kernel.
    }
    \vspace{-1em}
     \label{fig:model_overview}
\end{figure}

\section{Introduction}
\label{sec:intro}
Generative sequential models such as Large Language Models (LLMs)~\citep{radford2018improving,achiam2023gpt,bai2023qwen,grattafiori2024llama,guo2025deepseek} have typically been trained autoregressively, due to the ease of formulating a generative task as a next-token prediction problem, and parallelizing the learning of its conditional probability distribution. 

Autoregressive modeling is most effective when the order in which the data is presented is aligned with the direction of causality. However, not all tasks align well with autoregressive generation. 
For example, LLMs struggle to generate a large document with self-consistent reference sections, chapters, and word count for blocks of text. These references and numbers should be determined after the majority of the text has been generated, but because of the autoregressive constraint, the model may have to choose a section number arbitrarily and then later have to force the word count or numbers of sections to fit the requirement.
A related challenge is the difficulty of modifying or improving generated outputs. A useful ability for a generative model would be to insert, edit, or delete text with ease, as one would do with a text editor. This would also allow for a more hierarchical generation, where the model can first output an outline of the solution as some form of high-level blueprint, summary, or bullet points, and then edit and refine the details to obtain the final solution.

This motivates a decoding mechanism that is more flexible and allows amendments in hindsight. A recent surge in work on discrete diffusion~\citep{austin2021structured,lou2023discrete,nie2025large} attempts to address this shortcoming as an iterative token demasking and correction task. 
However, there are several limitations to this approach. First, most diffusion architectures generate a fixed-length output, which is not effective for applications such as chat systems, where model generations are of arbitrary length. Second, discrete diffusion regenerates the entire text using a bi-directional encoder, which cannot reuse previous computations with KV caching, making inference less efficient. Third, even though the token-correction mechanism could allow modifications to previously generated output, having a fixed position for each token prevents insertion or deletion of arbitrary numbers of tokens. Block diffusion~\citep{arriola2025block} combines autoregressive generation with block-wise discrete diffusion, alleviating the first two issues. However, such approaches are still constrained by the absolute positions at which tokens were generated.

We postulate that the limiting factor in existing work is that they do not distinguish between the \textit{generated positions} and the \textit{display positions} of tokens when they apply positional encoding. By decoupling these two positions, the model can generate tokens in a flexible order that is then parsed and reordered into the display order. In this work, we propose a novel training paradigm, which applies a set of inverse edit operations to the target sequence, and train the model to undo the inverse edit operations, effectively making the model learn to edit and refine the context that it generated. This is analogous to diffusion training, in which the model is trained to undo a noising operation. 

The uniqueness of our method is that the model reassigns a new set of positional encodings to tokens in the context according to the \textit{display order} during and after each edit, as opposed to the \textit{generated order}. The model can perform operations such as insertion, editing, and deletion to manipulate existing outputs in a similar way to how we write documents using a text editor. Once tokens are substituted or deleted, they are also masked out so that the model only sees the tokens currently in context in the \textit{display order} during further generation. Therefore, the underlying model is still autoregressive in its generation order, retaining its benefits such as KV-caching and early stopping of generation via an End-of-Sequence token \eostoken, but can be non-autoregressive in the display order. 

We name the method AutoIndexer because it learns to propose the positions in the context for editing. AutoIndexer has a marker head to determine the timing of the edit, and a cursor head that predicts the edit cursor positions, in addition to the vocabulary token head. This is in contrast to Fill-in-the-Middle (FIM)~\citep{bavarian2022efficient}, where the section to be edited is given, both during training and at inference.

\begin{center}
\begin{qualbox}
\qualrow{RAW}{\texttt{<h4 className="col-12 col-md-12 my-3 mx-auto contact-t \corr{\ldots} question or are you a hairstylist looking to join us?</h4>\ldots\editmk{61,61}itles">Do you have a\returnmk\ldots}}
\qualrow{PARSED}{\texttt{<h4 className="col-12 col-md-12 my-3 mx-auto contact-titles">Do you have a question or are you a hairstylist looking to join us?</h4>}}
\end{qualbox}
\vspace{0.5em}
\captionof{figure}{Example of a successful edit by AutoIndexer. It opens an edit at position \texttt{[61, 61)}, inserts the missing text, and returns to continue generation. \corr{\ldots} is where the omission of the prompt took place (omission location is not disclosed to the model). More examples in \Cref{app:edit_qualitative}.}
\label{fig:intro-edit-example}
\vspace{-1em}
\end{center}

\begin{table}[!ht]
    \caption{Comparison between decoding strategies of transformers.}
    \label{tab:decoding-strategies}
    \centering
    \small
    \begin{tabular}{lccc}
        \toprule
        & Autoregressive & Discrete Diffusion & Chain of Edits (Ours) \\
        \midrule
        Generation order of tokens    & Autoregressive & Non-autoregressive & Autoregressive \\
        Display order of tokens   & Fixed-order & Fixed-order & Decoupled (flexible) \\
        Prediction target   & Next token & Token demasking & Undo perturbation \\
        KV-caching   & Exact & Approximate & Exact \\
        Amendable context   & No & Limited & Yes \\
        \bottomrule
    \end{tabular}
    \vspace{-1em}
\end{table}

Our contributions are the following: (a) \textbf{Active Context Refinement:} a model that can modify its output in any order without length constraints, by making it decide when and where an edit is required and move a cursor to any position in the sequence, unlike Fill-in-the-Middle~\citep{bavarian2022efficient} where the edit span is given externally; (b) \textbf{Perturbation training:} a self-supervised learning paradigm to teach the model to edit by perturbing each ground-truth sequence with randomly sampled inverse edit operations, and training the model to autoregressively undo this perturbation, analogous to a diffusion objective but over discrete edit inverses rather than noise; (c) \textbf{Fused Kernel with Chain-of-Edits RoPE:} an attention kernel that computes the relative positional encodings using RoPE~\citep{su2024roformer} but for multiple edits in one forward pass for training efficiency, informing the model of the updated token positions after each edit operation; and (d) \textbf{Sequence Parser:} a parser that converts the model's generation with multiple edit operations back to \textit{display order}.

\section{AutoIndexer}

AutoIndexer's decoding mechanism decouples generation order and display order. While reusing the transformer backbone, AutoIndexer augments conventional next-token prediction with a redesigned model training method, a cursor prediction head, edit-aware positional encoding, and a parser that maps the raw generation to a display sequence. The overview of the method is shown in \Cref{fig:model_overview}.

\subsection{Active Context Refinement via marker tokens and edit indices}
\label{sec:markers}

AutoIndexer is equipped with an additional marker head to initiate an edit, and a cursor head to define the position of the edit. The marker head can predict \normaltoken~for normal vocabulary tokens, \eostoken~for End-of-Sequence, \edittoken~or \returntoken, the latter two marking the start and end of an edit, analogous to how a cursor and selection work in a text editor.
The cursor head is a two-headed pointer network (an attention layer with just the query/key heads) on top of the final hidden states of the transformer backbone that predict an edit-start index $S$ and an edit-end index $E$. The edit starts and ends immediately \textit{before} the tokens at indices $S$ and $E$.
While $S$ and $E$ are defined in the \textit{generated order}, the edit indices $s$ and $e$ in the corresponding current \textit{display} sequence are guaranteed to satisfy
$s \le e$.
This single operation covers insertion ($s = e$), deletion ($s < e$ with no replacement tokens), and substitution ($s < e$ with replacement tokens).

The edit start cursor is sampled simultaneously when an \edittoken~is emitted, and the end cursor is sampled immediately following the \edittoken~(rather than at \returntoken). This defines the edit span (the tokens in the context that are selected for edits) so that the model can perform its edit with the end state in mind.
Note that unlike Fill-in-the-Middle~\citep{bavarian2022efficient}, where the edit span is supplied as input, the model learns to decide the location in the context to be edited.

\edittoken~moves the cursor to position $s$ within the current \textit{display} sequence, and shifts the positional encoding of existing tokens so that the newly generated token always has a relative position of 1. The \edittoken~token itself is initially positioned at the end of the current \textit{display} sequence, but is moved to the edit frontier once the edit starts, just after the tokens to be removed, as shown in \Cref{fig:model_overview}.

\returntoken~returns the cursor to the end of the current \textit{display} sequence and reassigns the post-edit positional encodings to tokens. This is when $e-s$ tokens are removed from the context.
Removed tokens are masked from future attention and omitted from the decoded output rather than physically deleted from the KV cache. This also allows hidden edit spans to serve as \textit{registers} that store temporary computation~\citep{goyal2023think,darcet2023vision,pfau2024let}, or as drafts and intermediate plans like Chain of Thought~\citep{wei2022chain} that are not a part of the final output.

\subsection{Training via simulated edit sequences}
\label{sec:perturbation}

At training time, AutoIndexer turns a clean sequence $\vx = (x_0, \dots, x_{T-1})$ into a perturbed target sequence by applying the \textit{inverse} of a randomly sampled edit program. We sample $n$ edit operators $\mathcal{E}_0, \dots, \mathcal{E}_{n-1}$ with randomly drawn (Poisson-distributed) insertion lengths, deletion lengths, cursor spans $[s_i, e_i)$, and post-edit append lengths.
Each edit replaces the context tokens in the range of $[s_i, e_i)$ in the display order with newly generated tokens in the \edittoken\ldots\returntoken~block, dropping the markers after completion. The operators are numbered in the order of generation, so $\mathcal{E}_0$ is the first edit and $\mathcal{E}_{n-1}$ produces the final output $\vx$.

An inverse edit operation $\mathcal{E}_i^{-1}$ undoes this by re-injecting the edit block.
Draft tokens that were edited out do not appear in the parsed output. We fill these positions with independently sampled valid tokens. These randomly sampled tokens are masked in the decoding target so the model learns to detect and remove erroneous context without learning to generate them.

\begin{figure}[!t]
     \centering
     \includegraphics[width=0.9\textwidth]{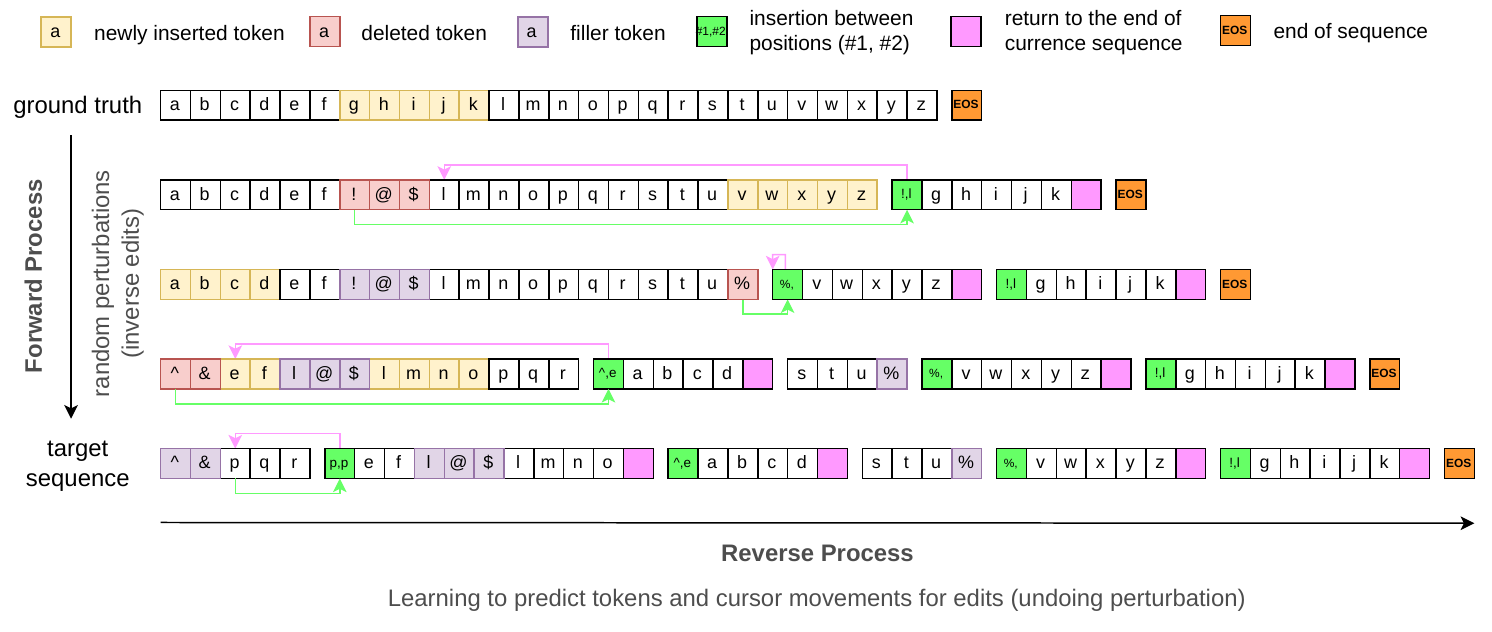}
    \caption{A simulated edit sequence for training. The ground-truth clean sequence (top) in display order is randomly perturbed to obtain a target sequence (bottom) with edit/return markers, which the model learns to generate. A sequence parser recovers the intended display order of the tokens.}
     \label{fig:target_sequences}
     \vspace{-1em}
\end{figure}

The target sequence $\tau$ is built by starting from the clean sequence $\vx$ in \textit{display} and undoing the edits one at a time, from last to first:
\begin{equation}
    \tau = \mathcal{E}_0^{-1} \circ \cdots \circ \mathcal{E}_{n-1}^{-1}(\vx).
    \label{eq:perturb}
\end{equation}
This procedure is described in \Cref{fig:target_sequences}. The way clean data are corrupted in the forward process and the model is trained to invert the corruption in the reverse process is analogous to the diffusion training objective, except that the corruption is a discrete edit program rather than masking or Gaussian noise. Since $\tau$ can be constructed from any sequential dataset, no human edit history is required, though real histories such as git diffs can be useful for task-specific fine-tuning.

\subsection{Position reassignment via Chain-of-Edits Rotary Position Encoding}
\label{sec:rope}

A technical challenge in training the model to perform multiple edits in one forward pass is how to handle positional encoding for tokens in the context. Since the model does not know the exact \textit{display order} of the tokens to be generated at inference time until all edits are completed, the token positions must be reassigned after each edit operation.
Emulating this positional shift over multiple edits for training is non-trivial, however, since the RoPE~\citep{su2024roformer} trick of factorizing the $T \times T$ relative position matrix into query/key rotations only works for static token sequences.

To solve this issue, we generalized RoPE to a \textit{Chain of Edits}, factorizing the attention computation into edit-wise segments whenever the token positions are shifted. Each segment corresponds to a contiguous token subsequence over which the token positions monotonically increase. The segment is either a normal segment (normal next token generation) or an in-edit segment containing the replacement tokens between \edittoken~and \returntoken. For a normal segment, a standard RoPE is applied. For an in-edit segment, as mentioned in \Cref{sec:markers}, \edittoken~signifies an ongoing edit and will be assigned a relative position of 1. All tokens preceding the edit cursor will have a normal RoPE applied, and all tokens succeeding the cursor will maintain the same position throughout the edit, meaning that the rotations are no longer relative to the query position for post-cursor tokens. 

For every segment, we compute and store (i) display positions $\vtheta_q$ for the query tokens in the block, (ii) display positions $\vtheta_k$ for all previously generated visible key tokens, (iii) an attention mask $m_{\mathrm{attn}}$ that excludes removed tokens and resolved edit/return markers, and (iv) a cursor-anchor mask $m_{\mathrm{anc}}$ for keys \textit{after} the cursor in the display order when an edit is underway. Where the cursor-anchor masks are applied, only the key is rotated and the query remains unrotated. The segment-wise attention is:
\begin{equation}
    \text{attn\_weight}(\vq, \vk, \vtheta_q, \vtheta_k, m_{\mathrm{anc}}, m_{\mathrm{attn}}) = (\vq_{\mathrm{rope}} \cdot \vk_{\mathrm{rope}} \odot \overline{m}_{\mathrm{anc}} + \vq \cdot \vk_{\mathrm{rope}} \odot m_{\mathrm{anc}}) \odot m_{\mathrm{attn}},
\end{equation}
where $\vq_{\mathrm{rope}}, \vk_{\mathrm{rope}}$ denote the RoPE-rotated query/key. The expression reduces to standard RoPE attention in the absence of edits, since $m_{\mathrm{anc}}$ will be false throughout.

By decoupling the token's fixed \textit{generation order} from its more flexible \textit{display order} via the edit mechanism, AutoIndexer enables exact KV-caching even when the token \textit{display order} is non-monotonic. Keys and values are cached once in unrotated form to allow KV cache reuse, and the rotary angle is applied when attention is computed. 

Although the edit operations are sequential, AutoIndexer does not train by replaying edits one at a time. Our custom fused attention kernel evaluates the complete packed sequence in one forward pass. It consumes the full query/key tensors, segment boundaries, and position tuples $(\vtheta_q, \vtheta_k, m_{\mathrm{anc}}, m_{\mathrm{attn}})$ per segment, and fuses RoPE, masking, dot product, softmax, dropout, and value aggregation without materializing the full attention matrix. More details about this kernel are described in \Cref{app:fa4}.

\subsection{Decoding and sequence parsing}
\label{sec:decoding}

At inference time, AutoIndexer samples a sequence $\hat\tau$ autoregressively, with KV-caching for both token and cursor heads. At each step, a dedicated marker head samples a marker class of \normaltoken, \edittoken, \returntoken, or \eostoken. Illegal classes are masked to $-\infty$. A normal vocabulary head sampling takes place when \normaltoken~is sampled. A decoding constraint enforces alternating \edittoken~and \returntoken~markers, so each edit is closed before another begins. When \edittoken~is emitted at display position $t$, the cursor head samples an edit-start index $s$ ($0 \leq s \leq t$), and an edit-end index $e$ ($s \leq e \leq t$) is sampled at the immediate following step (after the edit cursor and token positions have been updated). A sequence parser updates the token display-position mappings $\vtheta_q$ and $\vtheta_k$, and masks $m_{\mathrm{anc}}$ and $m_{\mathrm{attn}}$, as required in \Cref{sec:rope}. When \eostoken~or the maximum length is reached, the parser masks out deleted spans and emits the remaining tokens in display order.

The prediction is recovered by the sequence parser by iteratively applying edits as they are generated:
\begin{equation}
    \hat{\vx} = \mathcal{E}_{n-1} \circ \cdots \circ \mathcal{E}_0(\hat\tau).
    \label{eq:parse}
\end{equation}
Since $\mathcal{E}_i \circ \mathcal{E}_i^{-1} = \mathrm{id}$ by construction, parsing the synthetic sequence $\tau$ recovers $\vx$ exactly. 

\subsection{Training objective}
\label{sec:objective}
The perturbation program described in \Cref{sec:markers} yields both the perturbed target sequence $\tau$ and the target cursor positions ($S_i$, $E_i$) for each edit. Thus, both token prediction and edit localization are self-supervised.
The model's token head is trained on $\Ls_{\text{token}}$, a standard next-token cross-entropy over $\tau$, ignoring padding and masked tokens. The marker head is trained with a focal loss~\citep{lin2018focalloss} $\Ls_{\text{marker}} = -(1 - p_t)^\gamma \log(p_t)$, where $p_t$ is the predicted probability for the target marker class, to classify the marker class at each step, taking into account the class imbalance amongst \normaltoken, \edittoken, \returntoken, and \eostoken. The start- and end-cursor heads are trained on $\Ls_{\text{cursor}}$, a cross-entropy loss over targets $S_i$ and $E_i$ per edit, evaluated immediately before and at each \edittoken~marker, respectively.
The total training loss is $\Ls = \Ls_{\text{token}} + \Ls_{\text{marker}} + \Ls_{\text{cursor}}$. 
A stop-gradient mixing ratio controls how much the cursor loss updates the shared transformer backbone.
In practice, we found that it is better to detach the cursor head gradients entirely from the final hidden states to learn a better latent representation first, unaffected by the cursor objective.

Since AutoIndexer loads a standard transformer backbone, using the same attention layers apart from injecting the segment position tuples and the fused attention kernel in \Cref{sec:rope}, the backbone can be LoRA-finetuned~\citep{hu2022lora} from pre-trained models. Learnable components in addition to the backbone are the cursor head and the marker head.

\section{Experiments}

\input{tables/benchmarks_pre-training}

We evaluated AutoIndexer along two axes: retention of LLM capabilities in standard natural language and coding tasks, and the newly acquired error correction capabilities. 
First, we benchmarked AutoIndexer against autoregressive~\citep{yang2025qwen3} and diffusion~\citep
{nie2025large,ye2025dream} models on standard general, math/science, and code tasks in \cite{eval-harness}, both for base tasks and instruction tasks. 
Diffusion models tend to trade off prediction quality for the advantage of any-order decoding, due to KV-cache approximation. Benchmarking on standard LLM tasks allows us to test whether AutoIndexer maintains prediction quality, even with the added Chain-of-Edits capability. 
Next, we evaluated AutoIndexer's capability of active revision. Using the same base checkpoint used for LLM benchmarking, we ran a targeted edit diagnostic that tests whether the model can detect a corrupted span in the context, localize it, generate a suitable replacement, and resume generation coherently. Experimental details in \Cref{app:experimental_details}.

\paragraph{Pre-training benchmark results.}
\label{sec:pre-train_results}
\Cref{tab:benchmark-results} shows the base model evaluation results for general, math, science, and coding tasks.
AutoIndexer inherits the Qwen3 backbone's strong performance in downstream tasks even after conversion into a Chain-of-Edits model, consistently outperforming competing diffusion baselines for flexible-order decoding. The AutoIndexer model essentially tied with the matched Qwen3 LoRA baseline, suggesting that the introduction of marker and cursor heads, and token reordering due to Chain-of-Edits do not measurably affect performance compared to ordinary left-to-right decoding, in contrast to other diffusion baselines where degradation of performance is common.
Notably, AutoIndexer outperformed all other baselines by a wide margin for the HumanEval evaluation without instructions, where diffusion models especially struggled. This suggests that the edit-training objective transfers especially well to code, possibly because iteratively revising a partial solution resembles how code is naturally produced.

\paragraph{Post-training benchmark results.}
\Cref{tab:benchmark-results-sft} shows the evaluation results of supervised-fine-tuned models on instruction tasks. The results follow the same pattern of the AutoIndexer model outperforming the instruction-tuned diffusion baselines, and matching the Qwen3 8B LoRA baseline performance.
Interestingly, there seems to be a drop in performance of the LoRA-finetuned Qwen3 and AutoIndexer models compared to the original Qwen3 8B checkpoint for math/science and code tasks. General tasks such as ARC-Challenge and MMLU-Pro, on the other hand, retained their performance after LoRA-finetuning. We hypothesize that the Qwen3 instruction-finetuned checkpoint additionally benefited from Reinforcement Learning with Verifiable Rewards (RLVR), especially for maths and coding tasks, whereas we did not perform RLVR fine-tuning in the post-training stage. 
Nevertheless, the observation holds that our AutoIndexer model maintains superior instruction-following capabilities over the diffusion baselines.

\input{tables/benchmarks_post-training}

\paragraph{Edit-capability evaluation setup.}
To verify that AutoIndexer can actively revise its context during generation, we took the AutoIndexer 8B LoRA Base checkpoint obtained from CPT, and ran it against the held-out validation set of the CPT datamix (\Cref{app:data_mix}) with one corruption. We tested the model on insertion (by removing tokens from the ground truth), deletion (by adding noise tokens), substitution (by replacing tokens with noise), and no-op (the context was the ground truth). We inspected the checkpoint both under a teacher forcing setup (details and results reported in \Cref{app:teacher-forcing}) to measure the model's edit initiation and termination decisions, localization of the edit span (start and end cursor), and replacement token accuracy, and a free generation setup where the model was rolled out from a corrupted draft and a 64-token budget to test whether the model can fully execute edits on its own.

\begin{table}[b]
    \vspace{-1em}
    \caption{Free-generation per-outcome counts (100 examples per category, marker head sampled with the default settings of temperature $T_\text{mark}{=}1.0$ and marker top-$p{=}1.0$).}
    \label{tab:edit-free-generation}
    \centering
    \small
    \resizebox{\textwidth}{!}{%
    \begin{tabular}{lcccccc}
        \toprule
        Outcome / 100 & Start cursor & End cursor & Insert & Delete & Substitute & No modification \\
        \midrule
        Repaired exactly & \cmark & \cmark & 10 & 53 & 18 & -- \\
        No edit needed (true negative) & -- & -- & -- & -- & -- & 89 \\
        No edit emitted (false negative) & -- & -- & 57 & 10 & 8 & -- \\
        Correct span, improved & \cmark & \cmark & 12 & 9 & 29 & -- \\
        Correct span, neutral & \cmark & \cmark & 2 & 1 & 11 & -- \\
        Correct span, worse & \cmark & \cmark & 2 & 6 & 8 & -- \\
        Correct edit start, deleted too few & \cmark & \xmark & -- & 6 & 10 & -- \\
        Correct edit start, deleted too many & \cmark & \xmark & 4 & 4 & 6 & -- \\
        Incorrect position & \xmark & -- & 13 & 11 & 10 & 11 \\
        \midrule
        Correct edit emission & -- & -- & 43 & 90 & 92 & 89 \\
        Correct edit span & -- & -- & 26 & 69 & 66 & -- \\
        Net improvement & -- & -- & $+11$ & $+66$ & $+31$ & -- \\
        \bottomrule
    \end{tabular}%
    }
\end{table}

\paragraph{Edit-capability evaluation results.}

The free-generation results are shown in \Cref{tab:edit-free-generation}. The AutoIndexer model was able to correctly distinguish corrupt sequences from clean sequences with around $90\%$ accuracy for the delete, substitute, and no-op cases, and predicted the correct edit span for nearly $70\%$ of the delete and substitute cases, resulting in 53 exact repairs for deletion and 18 exact repairs for substitution. ``Net improvement'' is the number of edits that reduced the Levenshtein distance to the ground truth. Correction via insertion remains the most challenging task, since the lack of tokens is easy to miss. Less than half of corrupted drafts ($43\%$) triggered an edit, but when an edit occurred, it localized the insertion correctly $60\%$ of the time. Qualitative results in \Cref{app:edit_qualitative} indicate that AutoIndexer employs a semantic and syntactic understanding of the context to apply informed edits, rather than randomly modifying its context. Visualization of the attention weights (\Cref{fig:attention_weights}) and cursor head (\Cref{fig:cursor_pointer_network}) show that the model is able to localize erroneous content and focus on the surrounding context.

\section{Related work}
\paragraph{Transformers} Transformers~\citep{vaswani2017attention} are a versatile architecture family for generative modeling, widely adopted across numerous application domains~\citep{achiam2023gpt,li2022competition,dosovitskiy2020image,radford2023robust,jumper2021highly,hu2023gaia,ishida2024langprop,black2024pi_0,kanervisto2025world}. While full attention encoder models have been effectively used to extract context~\citep{vaswani2017attention}, learn feature representations~\citep{devlin2019bert}, or generate fixed-size outputs such as images~\citep{peebles2023scalable}, autoregressive decoder models have been more widely adopted for generative modeling in sequential domains such as language, time series, and videos, because they model the joint probability of the outputs more accurately and can produce outputs of arbitrary length.

\paragraph{Discrete Diffusion} There have been efforts to apply continuous diffusion to LLMs by learning a mapping between discrete and continuous spaces~\citep{li2022diffusion,dieleman2022continuous,lovelace2023latent,jo2025continuous}, or to adapt the diffusion formulation for discrete spaces~\citep{austin2021structured,campbell2022continuous,lou2023discrete}. Alternatively, several works~\citep{sahoo2024simple,shi2024simplified,nie2025large} have proposed a Masked Diffusion Model approach that iteratively replaces token masks with token predictions. While such approaches are easy to implement and can even be adapted from pre-trained autoregressive models~\citep{gong2025scaling}, they share the inherent limitation that the models can only generate a fixed-length output, and that tokens are positionally constrained once generated, with limited capabilities of context refinement.

Block Diffusion~\citep{arriola2025block} alleviates the first limitation by retaining a block-wise autoregressive structure while applying masked diffusion within each block using a full attention model, thereby trying to achieve the best of both worlds of using full attention blocks (i.e. parallelized token generation leading to speed up in generation) and an autoregressive mechanism (i.e. flexible output length and being able to take advantage of KV caching during inference for efficiency).

\paragraph{Any-order autoregressive generation} Building on permutation language models~\citep{yang2019xlnet}, recent work revives any-order autoregressive factorizations to match diffusion's flexibility. Any-subset speculative decoding predicts and verifies multiple positions per step~\citep{guo2025reviving}, and A3~\citep{du2026autoregressive} extends this to arbitrary token groups via two-stream attention. Unlike AutoIndexer, both still commit upfront to a fixed set of positions to fill, rather than growing or shrinking the sequence through edits.

\paragraph{Edit-based sequence generation} Classic insertion/deletion decoders~\citep{stern2019insertion,gu2019levenshtein,chan2019kermit} and their diffusion counterpart, DiffusER~\citep{reid2023diffuser}, have recently been generalized further: Edit Flows~\citep{havasi2025edit} casts edit-based generation as a continuous-time Markov chain trained via flow-matching, and Beyond Masks~\citep{ding2026beyond} makes deletion/insertion the diffusion process itself, removing mask and pad tokens entirely. All of these, however, remain non-autoregressive and cannot reuse a KV cache, since token positions shift between steps. Two concurrent works keep edits strictly autoregressive: Reviser~\citep{diab2026reviser} predicts cursor actions (insert, move, stop) over a mutable canvas, and Stream of Revision~\citep{yang2026stream} lets a code model backtrack via dedicated action tokens. Our method differs in that it uses a pointer network to localize edits, and adjusts the positional encoding after each edit, enabling continuous revision and extension without re-computing the KV.

\section{Conclusion}

AutoIndexer enhances autoregressive models with active context editing and refinement capabilities. It decouples the generation order of tokens and the displayed order of tokens, allowing the model to revisit earlier drafts and iteratively refine the output sequence. The experiments suggest that this edit capability can be introduced while preserving the broad capabilities of the autoregressive backbone. Unlike other editing methods such as tool-calling and Fill-in-the-Middle, the model and training strategy are architected to account for context editing, such that the attention implements edit-aware positional encodings, and the KV-cache at inference exactly matches the training distribution, enabling the model to perform multiple edits without re-computing previous forward passes.

Chain-of-Edits decoding is especially useful in domains where consistency and internal cross-referencing are solidified over iterative refinement, including code (e.g. finetuning the model directly on git diffs), long-form writing, mathematical derivations, and structured documents. It could also be compatible with drafting and planning, commonly seen in Chain-of-Thought and tool calling, where intermediate thought or tool call traces can be discarded before the final output. AutoIndexer is an initial demonstration of causal transformers with continual context editing capabilities, suggesting a path toward generative systems that have the entire loop of planning, self-correction, and in-place refinement of their outputs built into their optimization objective.

\bibliographystyle{unsrt}
\bibliography{main}

\newpage

\crefalias{section}{appendix}%
\crefalias{subsection}{appendix}%
\crefalias{subsubsection}{appendix}
\appendix

\section*{Reproducibility statement}
The code that implements the AutoIndexer method along with the fused accelerated attention kernels and training scripts are available at \url{https://github.com/AutodeskAILab/autoindexer}.
Details on the training datasets and evaluation frameworks are available in the Appendix.

\section{Experimental Details}
\label{app:experimental_details}

\begin{figure}[!b]
     \centering
     \includegraphics[width=\textwidth]{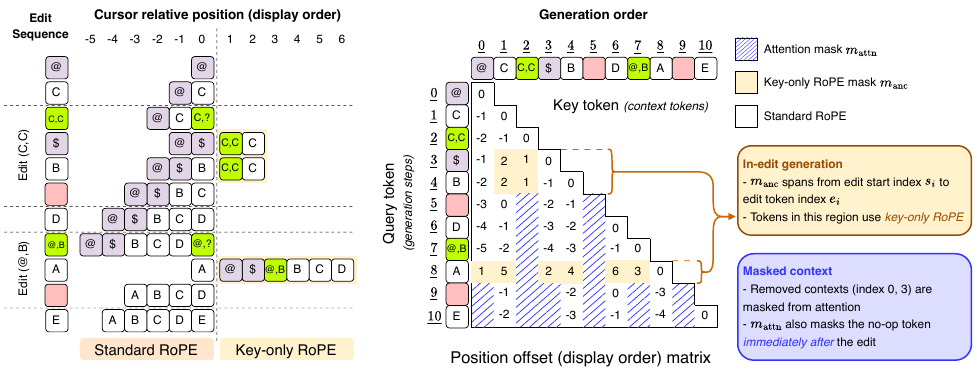}
    \caption{Relationship between generation order and display order of tokens. The left-hand-side shows the generated tokens taking the display order as reference, whereas the right-hand-side shows the relative positions of the tokens, taking the generation order as reference.
    } 
     \label{fig:gen_vs_display_order}
\end{figure}

\subsection{Mapping Chain-of-Edits Attention onto FlashAttention-4 Varlen}
\label{app:fa4}

A naive implementation of the routing rule in \Cref{sec:rope} materializes two full $T \times T$ attention score matrices per segment, $\vq_{\mathrm{rope}} \cdot \vk_{\mathrm{rope}}$ and $\vq \cdot \vk_{\mathrm{rope}}$, before $m_{\mathrm{anc}}$ selects which one each key uses -- consuming 2$\times$ the FLOPs of a conventional attention score matrix. Our optimized version instead routes the relative-attention path (attending only to keys scored with $\vq_{\mathrm{rope}}$) and the cursor-anchored attention path (attending only to keys scored with $\vq$) to two separate FlashAttention-4 \texttt{varlen} calls~\citep{zadouri2026flashattention4}, packing each path's keys into contiguous memory beforehand rather than masking a dense score matrix afterward. Since every key belongs to exactly one path, the two calls together score the same keys the naive implementation would, so the sum of FLOPs across both calls approximates 1$\times$ conventional attention rather than 2$\times$. The two paths' outputs are recombined with a log-sum-exp reweighting of their softmax statistics, which reconstructs the same joint softmax that \Cref{sec:rope}'s formula defines, so this restructuring changes only how the computation is organized, not the attention weights it produces. To reduce the overhead of the routing and packing logic, we implemented another kernel in CuTe-DSL~\citep{nvidia2025cutedsl}, fusing the routing logic with the rotation logic.

\subsection{Marker head decoding}
\label{app:edit_decoding}

At the inference step, AutoIndexer uses temperature $T_\text{voc}{=}0.7$, top-$p{=}0.8$, and top-$k{=}50$ to sample from the vocabulary token head, matching the Qwen3 baseline. Marker emission is a separate categorical decision from the marker head's softmax over \textsc{normal}, \textsc{edit}, \textsc{return}, and \textsc{eos}, with illegal classes masked to match alternating-edit constraints. Marker temperature and marker top-$p$ can be set independently of the vocabulary head. In the reported runs both default to $1.0$ (disabled). Before sampling, the \textsc{edit} logit is biased by $\lambda \log P_{\mathrm{start}}(\hat s) + b$ with $\lambda{=}1.0$ and $b=0.0$, where $\hat s$ is the start index sampled for that step. This is to ensure that edits with a confident edit localization are more likely to be triggered. Because markers are outside the vocabulary distribution, vocabulary top-$p$ and top-$k$ no longer affect whether an edit fires.

\subsection{Continued Pre-Training (CPT) Hyperparameters}
\label{app:hyperparameters}

Both matched LoRA~\citep{hu2022lora} fine-tuning runs start from the Qwen3 8B Base model and use bfloat16 weights. The fine-tuning stream contains 100K training samples truncated to 2048 tokens, corresponding to approximately 200M tokens before AutoIndexer perturbation expands the raw training sequence. The optimizer settings are AdamW-compatible Hugging Face trainer defaults with learning rate $10^{-4}$, weight decay 0.01, gradient clipping at 1.0, and 1000 warmup steps. The validation budget is 6400 samples, with evaluation and checkpointing every 1000 steps. Generation during training uses Qwen3 base-model sampling settings with temperature 0.7 and top-$p=0.8$.

For the matched Qwen3 LoRA baseline, we use rank $r=16$, LoRA $\alpha=64$, dropout 0.05, all linear modules as targets, no bias terms, and effective training batch size 8. For AutoIndexer 8B LoRA, we use the same rank, alpha, dropout, target-module, and bias choices. In addition to the LoRA adapter weights, AutoIndexer saves the cursor head and marker head, which are specific to edit localization and marker decisions.

We used eight NVIDIA H100 GPUs with 80 GB memory for fine-tuning with DeepSpeed ZeRO-2. Under that setting, the Qwen3 LoRA and AutoIndexer LoRA runs have an effective batch of 8 samples per optimizer step and therefore train for about 12,500 steps over 100K samples. 

The AutoIndexer benchmark configuration uses perturbation probability 0.5, so half of training examples are converted into Chain-of-Edits sequences and the remainder are ordinary causal examples. The Chain-of-Edits model uses maximum raw sequence length 3072 to leave room for marker tokens and inserted content beyond the 2048-token clean examples. Synthetic edit sequences use mean number of edits 50, mean insertion length 5, mean deletion length 5, and mean extension length 40, with edits drawn uniformly at random from insertion, deletion, and substitution (a 1:1:1 ratio) rather than a fixed pure-edit fraction. Half of the edits are sampled with a short lag, with mean lag 5, and half remain uniformly placed to preserve long-range cursor supervision. The hidden state is detached before being passed to the cursor head, preventing the cursor head gradients from back-propagating through the shared backbone representation. The marker prediction loss uses a focal loss with focal parameter $\gamma=1.0$ and weight 1.0.

\subsection{Continued Pre-Training (CPT) Data Mix}
\label{app:data_mix}

\begin{table}[t!]
    \caption{CPT data mix used for the matched Qwen3 LoRA and AutoIndexer LoRA runs reported in \Cref{tab:benchmark-results}.}
    \label{tab:datamix}
    \centering
    \begin{tabular}{llr}
        \toprule
        Source & Dataset / subset & Weight \\
        \midrule
        DCLM-baseline & \texttt{mlfoundations/dclm-baseline-1.0-parquet} & 30 \\
        FineWeb-Edu & \texttt{HuggingFaceFW/fineweb-edu} & 15 \\
        SlimPajama & \texttt{AlppAI/SlimPajama-chunked} & 25 \\
        OpenWebMath & \texttt{open-web-math/open-web-math} & 5 \\
        StarCoderData Python & \texttt{bigcode/starcoderdata/python} & 7 \\
        StarCoderData JavaScript & \texttt{bigcode/starcoderdata/javascript} & 4 \\
        StarCoderData Java & \texttt{bigcode/starcoderdata/java} & 4 \\
        StarCoderData C++ & \texttt{bigcode/starcoderdata/cpp} & 3 \\
        StarCoderData TypeScript & \texttt{bigcode/starcoderdata/typescript} & 2 \\
        StarCoderData C & \texttt{bigcode/starcoderdata/c} & 2 \\
        StarCoderData Go & \texttt{bigcode/starcoderdata/go} & 2 \\
        StarCoderData Shell & \texttt{bigcode/starcoderdata/shell} & 1 \\
        \bottomrule
    \end{tabular}
\end{table}

The fine-tuning stream is an interleaved Hugging Face dataset mix. Weights are normalized to probabilities by the data loader, so the values below also give the approximate percentages used by the 100K-sample run. The composition is shown in \Cref{tab:datamix}.

In aggregate, the mix assigns 30\% weight to DCLM web text, 15\% to educational web text, 25\% to SlimPajama's diversified web/books/papers/Wikipedia/StackExchange mixture, 5\% to math-heavy OpenWebMath, and 25\% to source code from StarCoderData. The code portion is intentionally uneven across languages, with the largest share assigned to Python and smaller but nonzero shares for JavaScript, Java, C++, TypeScript, C, Go, and shell.

\subsection{Supervised Fine-Tuning (SFT) Hyperparameters}
\label{app:sft_hyperparameters}

The post-training results in \Cref{tab:benchmark-results-sft} use the same LoRA~\citep{hu2022lora} configuration, optimizer settings, and hardware as the CPT runs in \Cref{app:hyperparameters} (rank $r=16$, LoRA $\alpha=64$, dropout 0.05, all linear modules as targets, no bias terms, learning rate $10^{-4}$, weight decay 0.01, gradient clipping at 1.0, 1000 warmup steps, bfloat16 weights, eight NVIDIA H100 GPUs with DeepSpeed), but start from the instruction-tuned Qwen3 8B checkpoint~\citep{yang2025qwen3} rather than Qwen3 8B Base, and are trained on the instruction mix in \Cref{app:sft_data_mix} instead of the CPT mix. As with the CPT runs, responses are rendered through the chat template with thinking disabled, matching how the same tasks are scored at evaluation time. The AutoIndexer model configuration (perturbation probability, mean edit counts, edit-lag sampling, cursor head gradient ratio) is unchanged from \Cref{app:hyperparameters}, except that edits are restricted to the response portion of each example so that the prompt is never perturbed, mirroring how the prompt is otherwise kept intact during instruction fine-tuning.

\subsection{Supervised Fine-Tuning (SFT) Data Mix}
\label{app:sft_data_mix}

\begin{table}[t!]
    \caption{SFT instruction data mix used for the matched Qwen3 LoRA and AutoIndexer LoRA runs reported in \Cref{tab:benchmark-results-sft}.}
    \label{tab:sft-datamix}
    \centering
    \resizebox{\textwidth}{!}{%
    \begin{tabular}{llr}
        \toprule
        Source & Dataset / subset & Weight \\
        \midrule
        \multicolumn{3}{l}{\textit{General-purpose instructions}} \\
        Alpaca & \texttt{tatsu-lab/alpaca} & 15 \\
        Dolly-15k & \texttt{databricks/databricks-dolly-15k} & 10 \\
        WizardLM Evol-Instruct-70k & \texttt{WizardLMTeam/WizardLM\_evol\_instruct\_70k} & 20 \\
        UltraChat-200k (multi-turn) & \texttt{HuggingFaceH4/ultrachat\_200k} & 25 \\
        \midrule
        \multicolumn{3}{l}{\textit{Coding instructions}} \\
        CodeAlpaca-20k & \texttt{sahil2801/CodeAlpaca-20k} & 10 \\
        Evol-Instruct-Code-80k & \texttt{nickrosh/Evol-Instruct-Code-80k-v1} & 20 \\
        Magicoder OSS-Instruct-75k & \texttt{ise-uiuc/Magicoder-OSS-Instruct-75K} & 20 \\
        Glaive-code-assistant & \texttt{glaiveai/glaive-code-assistant-v3} & 20 \\
        \midrule
        \multicolumn{3}{l}{\textit{Math instructions}} \\
        MetaMathQA & \texttt{meta-math/MetaMathQA} & 20 \\
        \midrule
        \multicolumn{3}{l}{\textit{General-capability replay}} \\
        DCLM-baseline (replay) & \texttt{mlfoundations/dclm-baseline-1.0-parquet} & 10 \\
        StarCoderData Python (replay) & \texttt{bigcode/starcoderdata/python} & 10 \\
        \bottomrule
    \end{tabular}%
    }
\end{table}

As with the CPT mix in \Cref{app:data_mix}, the SFT stream in \Cref{tab:sft-datamix} is an interleaved Hugging Face dataset mix with weights normalized to probabilities by the data loader. The mix combines eight instruction sources spanning general-purpose and coding instructions (Alpaca, Dolly-15k, WizardLM Evol-Instruct-70k, UltraChat-200k, CodeAlpaca-20k, Evol-Instruct-Code-80k, Magicoder OSS-Instruct-75k, and Glaive-code-assistant) with math instructions (MetaMathQA) added because the original eight sources contain no math data. In aggregate, roughly 37\% of the mix is general-purpose instructions, 37\% is coding instructions, and 11\% is math instructions.

The remaining $\sim$17\% of the mix addresses a regression observed after LoRA SFT on only the instruction sources above: few-shot-scored evaluation tasks such as GSM8K (CoT) and MBPP (instruct) are scored by packing several solved exemplars and the real question into a single turn, a shape none of the instruction sources reproduce, since each is a single instruction with no in-context exemplars. Two few-shot-packed replay sources reproduce this exact shape at the same shot counts used at evaluation time (5-shot for GSM8K, 3-shot for MBPP; $\sim$9\% combined). Two general-capability replay sources reuse raw CPT-style text (the same underlying corpora as \Cref{app:data_mix}, restricted to DCLM web text and StarCoderData Python) trained with unmasked full-sequence loss rather than an instruction/response split, to keep some pressure on general and few-shot ability that a purely instruction-formatted mix does not otherwise reinforce ($\sim$7\% combined).

\paragraph{LLM benchmark protocol.}
All results were obtained with the dLLM~\citep{zhou2026dllm} evaluation stack, built on the LM Evaluation Harness~\citep{eval-harness}. We trained and evaluated two checkpoints, one for the base tasks and one for the instruction tasks. AutoIndexer 8B LoRA Base was LoRA-finetuned~\citep{hu2022lora} from the Qwen3 8B Base~\citep{yang2025qwen3} checkpoint, while AutoIndexer 8B LoRA Instruct was finetuned from the instruction-finetuned Qwen3 8B checkpoint. 

For the pre-training evaluation, we used base-model tasks from MMLU, ARC-Challenge, HellaSwag, WinoGrande, PIQA, GSM8K, MATH500, HumanEval, and MBPP. Generation was capped at 1024 tokens. The diffusion baselines were LLaDA 8B Base~\citep{nie2025large} and Dream 7B Base~\citep{ye2025dream}, and the autoregressive baseline was Qwen3 8B Base. We additionally included a Qwen3 8B LoRA model trained on the same Continued Pre-Training (CPT) datamix (\Cref{app:data_mix}) as AutoIndexer 8B LoRA Base to control for the effect of fine-tuning itself. AutoIndexer's outputs were scored after parsing its generated edit stream back into display order.
For the post-training evaluation, we used instruction-formatted task variants from ARC-Challenge, MMLU-Pro, GPQA-Diamond, GSM8K, MATH500, HumanEval, and MBPP, with responses rendered through each model's chat template. Generation was capped at 512 tokens. The diffusion baselines were LLaDA 8B Instruct and Dream 7B Instruct, and the autoregressive baseline was the instruction-tuned Qwen3 8B checkpoint. The matched Qwen3 8B LoRA and AutoIndexer 8B LoRA models were further fine-tuned from the Qwen3 8B checkpoint on the Supervised Fine-Tuning (SFT) instruction mix (\Cref{app:sft_data_mix}).

\subsection{Edit-Capability Diagnostic}
\label{app:edit_diagnostic}

The edit diagnostic evaluates the same AutoIndexer 8B LoRA model checkpoint reported in \Cref{tab:benchmark-results} on the validation set of the data mix in \Cref{app:data_mix}. Teacher-forced and free-generation runs each use 100 windows per operation, with an average edit size of $k{=}5$ and one 64-token rollout per window for free generation.

\subsubsection{Teacher-forced edit evaluation}
\label{app:teacher-forcing}

\Cref{tab:edit-teacher-forced} shows the \edittoken~and \returntoken~emission probabilities and the start and end cursor locations (edit span) accuracy in a teacher-forced setting. The model detects an incoherency in the text and correctly assigns a higher edit probability to the corrupt sequence than the clean sequence in almost all cases. The edit emission probability at the exact timestep as the ground truth is low because of an arbitrary lag between an error and its correction.
The model also identifies the edit spans for substitution and deletion with around $70\%$ accuracy. The weaker insertion signal is expected, since missing text leaves no explicit cue, unlike for substitution and deletion cases where noise tokens are randomly sampled.

\input{tables/teacher-forced-edits}

\subsubsection{Qualitative examples}
\label{app:edit_qualitative}

The examples below are drawn from free-generation rollouts under the decoding defaults in \Cref{app:edit_decoding}. Each card's header shows an outcome tag (\successtag{Exact repair}, \partialtag{Partial repair}, or \failuretag{Failure}) and the Levenshtein distance to ground truth before $\to$ after the edit. Corrupted spans are shown as \corr{\ldots}, with the actual corrupted tokens shown inside for the delete and substitute cases (the tokens a correct edit must remove). \texttt{[?]} stands in for tokens outside the Latin-1 range (e.g., CJK, Thai, Hebrew, emoji), which the typeset font cannot render. \editmk{$s$,$e$} marks the opening of an edit at cursor positions $s,e$ in the display sequence, and \returnmk\ marks the return of the cursor to the end of the stream.

\paragraph{Exact repair --- pure insertion (10/100).}
Insertion triggers least often of the three operations (43/100), but when it fires close to the omission it can still repair exactly. Six words are missing from an HTML heading. The model localizes the empty span correctly---\texttt{[61, 61)}---and reconstructs the missing text verbatim.

\begin{qualbox}
\qualhead{\successtag{Exact repair}}{\quallev{6}{0}}
\qualrow{ORIGINAL}{\texttt{<h4 className="col-12 col-md-12 my-3 mx-auto contact-titles">Do you have a question or are you a hairstylist looking to join us?</h4>}}
\qualrow{DRAFT}{\texttt{<h4 className="col-12 col-md-12 my-3 mx-auto contact-t \corr{\ldots} question or are you a hairstylist looking to join us?</h4>}}
\qualrow{EMITTED}{\texttt{\ldots className="col-12 col-md-12\editmk{61,61}itles">Do you have a\returnmk\ mx-auto">\ldots}}
\qualrow{RESOLVED}{\texttt{<h4 className="col-12 col-md-12 my-3 mx-auto contact-titles">Do you have a question or are you a hairstylist looking to join us?</h4>}}
\end{qualbox}

\paragraph{Exact repair --- pure deletion (53/100).}
Five garbage tokens were inserted into a C++ command-line option string. The model continues past the corruption, then reaches back and deletes exactly the injected span with no replacement:

\begin{qualbox}
\qualhead{\successtag{Exact repair}}{\quallev{5}{0}}
\qualrow{ORIGINAL}{\texttt{auto mask\_img\_path = op.add<popl::Value<std::string>>("", "mask", "mask image path", "");}}
\qualrow{DRAFT}{\texttt{auto mask\_img\_path = op.add<popl::Value<std::string>>("", "mask", "mask image \corr{[?] soaking principalmente} path", "");}}
\qualrow{EMITTED}{\texttt{\ldots::Switch \editmk{129,134} \returnmk\ >("s", "no-sleep", "do not sleep after processing");\ldots}}
\qualrow{RESOLVED}{\texttt{auto mask\_img\_path = op.add<popl::Value<std::string>>("", "mask", "mask image path", "");}}
\end{qualbox}

\paragraph{Exact repair --- substitution (18/100).}
Three garbage tokens replaced part of a Java import statement. The model locates the corrupted span and rewrites it to match the intended package path exactly:

\begin{qualbox}
\qualhead{\successtag{Exact repair}}{\quallev{3}{0}}
\qualrow{ORIGINAL}{\texttt{import org.netbeans.modules.editor.*; import org.netbeans.modules.lsp.client.LSPBindings; import org.netbeans.modules.lsp.client.Utils;}}
\qualrow{DRAFT}{\texttt{import org.netbeans.modules.editor.*; import org.netbeans \corr{enco(keys brutal} .client.LSPBindings; import org.netbeans.modules.lsp.client.Utils;}}
\qualrow{EMITTED}{\texttt{\ldots modules.editor.*; \editmk{185,188} .modules.lsp \returnmk\ import org.openide.text.Line;\ldots}}
\qualrow{RESOLVED}{\texttt{import org.netbeans.modules.editor.*; import org.netbeans.modules.lsp.client.LSPBindings; import org.netbeans.modules.lsp.client.Utils;}}
\end{qualbox}

\paragraph{Correct abstention --- no modification (89/100).}
On most uncorrupted drafts the model continues fluently and terminates without ever emitting a marker:

\begin{qualbox}
\qualhead{\successtag{Abstained}}{\quallev{0}{0}}
\qualrow{ORIGINAL}{(this window is uncorrupted, so it is identical to the draft below)}
\qualrow{DRAFT}{\texttt{\{ 1, 2 \} >= \{ 1, 2 \} true Every set is a superset of itself.}}
\qualrow{EMITTED}{\texttt{\ldots (no marker emitted)\ldots}}
\qualrow{RESOLVED}{(unmodified)}
\end{qualbox}

\paragraph{Partial repair --- pure deletion (9/100).}
Six garbage tokens were inserted into the middle of ``Embassies''. The model finds the exact span but, instead of deleting it cleanly, writes a short unwanted continuation in its place:

\begin{qualbox}
\qualhead{\partialtag{Partial repair}}{\quallev{6}{4}}
\qualrow{ORIGINAL}{\texttt{network of counselor or representatives in French Embassies around the world (see joint map).}}
\qualrow{DRAFT}{\texttt{network of counselor or representatives in French Emb \corr{ educated[?] Identified[?] yup} assies around the world (see joint map).}}
\qualrow{EMITTED}{\texttt{\ldots responsible for: -\editmk{38,44}assies and Cons\returnmk\ issuing permits\ldots}}
\qualrow{RESOLVED}{\texttt{network of counselor or representatives in French Embassies and Consassies around the world (see joint map).}}
\end{qualbox}

\paragraph{Partial repair --- substitution (29/100).}
Three garbage tokens replaced part of a Go comment. The model finds the correct span but writes a shorter, malformed replacement, leaving a doubled word behind:

\begin{qualbox}
\qualhead{\partialtag{Partial repair}}{\quallev{3}{2}}
\qualrow{ORIGINAL}{\texttt{// Analysis is the method that kicks off all of the analysis plugins}}
\qualrow{DRAFT}{\texttt{// Analysis is the method that kicks \corr{bilità astronauts[?]} the analysis plugins}}
\qualrow{EMITTED}{\texttt{\ldots used to kick off the analysis\editmk{15,18} off the\returnmk\ of files as they are written\ldots}}
\qualrow{RESOLVED}{\texttt{// Analysis is the method that kicks off the the analysis plugins}}
\end{qualbox}

\paragraph{Partial repair --- pure insertion (12/100).}
More often, when insertion fires it localizes correctly but the content falls short. Here the model localizes a missing brace-and-assignment correctly---empty span \texttt{[38, 38)}---but writes a shorter replacement than the gold six-token insertion:

\begin{qualbox}
\qualhead{\partialtag{Partial repair}}{\quallev{6}{5}}
\qualrow{ORIGINAL}{\texttt{if(!transpose) \{ start = -1; step = 1; \} else \{ start = k; step = -1; \} \} else \{ \ldots\}}}
\qualrow{DRAFT}{\texttt{if(!transpose) \{ start = -1; step = 1; \} else \corr{\ldots} step = -1; \} \} else \{ \ldots\}}}
\qualrow{EMITTED}{\texttt{\ldots while(j != -1) \editmk{38,38} \{ \returnmk\ \{ q[i][j] = 0; j += step; \} \}\ldots}}
\qualrow{RESOLVED}{\texttt{if(!transpose) \{ start = -1; step = 1; \} else \{ step = -1; \} \} else \{ \ldots\}}}
\end{qualbox}

\paragraph{Failure --- wrong position for pure deletion (11/100).}
Six garbage tokens were inserted mid-expression in a C++ file, but the model's cursor lands one token late. The first corrupted token survives as an orphaned word rather than being removed:

\begin{qualbox}
\qualhead{\failuretag{Failure}}{\quallev{6}{1}}
\qualrow{ORIGINAL}{\texttt{auto \&scores\_dim = scores->dims();}}
\qualrow{DRAFT}{\texttt{auto \&scores\_dim = \corr{ mutated Edit\_Max.related):?> fullName} scores->dims();}}
\qualrow{EMITTED}{\texttt{\ldots float *r\editmk{218,223}\returnmk pn\_rois\_data = rpn\_rois->data<float>();\ldots}}
\qualrow{RESOLVED}{\texttt{auto \&scores\_dim = mutated scores->dims();}}
\end{qualbox}

\paragraph{Failure --- wrong position for insertion (13/100).}
Two characters are missing from a hex color literal at position 67, but the edit does not fire until position 219, 152 tokens later, right at the end of the draft. The empty-content marker pair fires back-to-back and inserts nothing, leaving the gap unfilled:

\begin{qualbox}
\qualhead{\failuretag{Failure}}{\quallev{2}{2}}
\qualrow{ORIGINAL}{\texttt{gray: '\#949BA6', lightGray: "\#e8e8e9",}}
\qualrow{DRAFT}{\texttt{gray: '\#949 \corr{\ldots} ', lightGray: "\#e8e8e9",}}
\qualrow{EMITTED}{\texttt{\ldots export const theme = createTheme \editmk{219,219} \returnmk\ (colors, fonts) [EOS]}}
\qualrow{RESOLVED}{\texttt{gray: '\#949', lightGray: "\#e8e8e9",}}
\end{qualbox}

\paragraph{Failure --- wrong position for substitution (10/100).}
Twelve garbage tokens replaced a Java method-chain expression at position 42, but the model's oracle-free cursor lands one token late and cuts a shorter span (\texttt{[43, 49)} instead of the gold \texttt{[42, 54)}), leaving half of the garbage in place and stitching in an unrelated method call:

\begin{qualbox}
\qualhead{\failuretag{Failure}}{\quallev{12}{12}}
\qualrow{ORIGINAL}{\texttt{selector().get(BoxSelectionAspect.class).boxEnabled = false; selector().size.set(type.size);}}
\qualrow{DRAFT}{\texttt{selector().get(BoxSelection \corr{.middle[?]hots Leonard[?]-product ACC genome,default azureTextWriter pred} .set(type.size);}}
\qualrow{EMITTED}{\texttt{\ldots getBlockTile(11, \editmk{43,49} ().\returnmk\ 3), orientation);\ldots}}
\qualrow{RESOLVED}{\texttt{selector().get(BoxSelection.middle(). genome,default azureTextWriter pred.set(type.size);}}
\end{qualbox}

\paragraph{Failure --- spurious edit on a clean draft (11/100).}
The draft is uncorrupted, but the model fires anyway partway through its own continuation, cutting into text it just wrote and replacing it with something unrelated to the source:

\begin{qualbox}
\qualhead{\failuretag{Failure}}{\quallev{0}{6}}
\qualrow{ORIGINAL}{(this window is uncorrupted, so it is identical to the draft below)}
\qualrow{DRAFT}{\texttt{\ldots less receptions but still WAY better numbers than Crabtree. Ever since the BCS was implemented, Strength of Schedule has become one of the biggest things in determining a team's grit and power.\ldots}}
\qualrow{EMITTED}{\texttt{\ldots biggest things in determining a team's \editmk{46,46} ranking. I think this is \returnmk\ It is a bad thing\ldots}}
\qualrow{RESOLVED}{\texttt{\ldots one of the biggest things in determining a team's ranking. I think this is grit and power.\ldots}}
\end{qualbox}

\newpage

\begin{figure}[!h]
     \centering
     \includegraphics[width=0.98\textwidth]{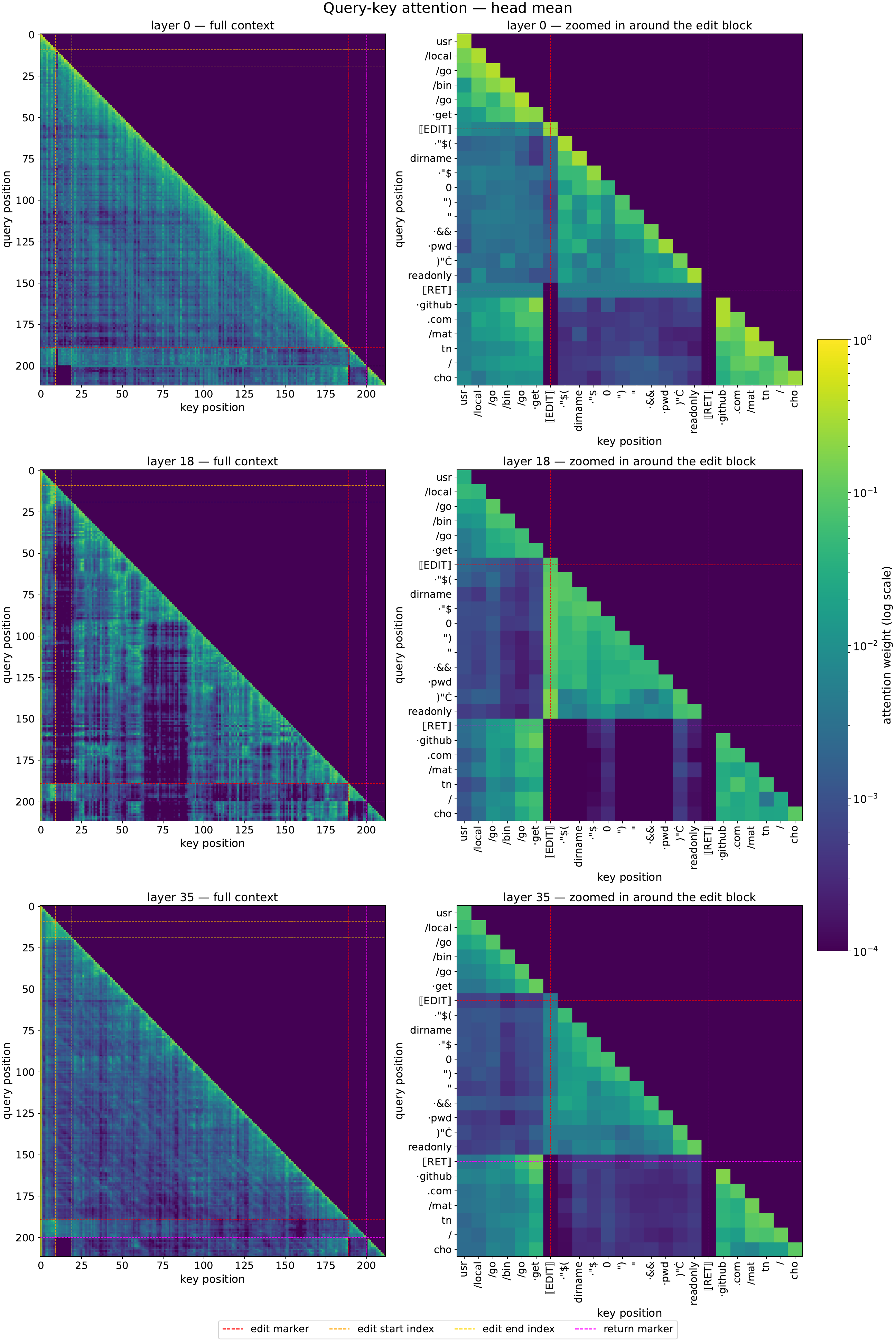}
    \caption{Visualization of attention weights averaged over the attention heads. The left-hand-side shows the full attention weights, while the right-hand-side shows a close-up around the edit block (\edittoken ... \returntoken). \edittoken\ and \returntoken\ are masked out upon the completion of the edit, as well as the deleted tokens. It can be seen that attention is focused before and after the erroneous token region during the edit so that the newly inserted tokens are coherent with the context.
    } 
     \label{fig:attention_weights}
\end{figure}

\begin{figure}[!h]
     \centering
     \includegraphics[width=\textwidth]{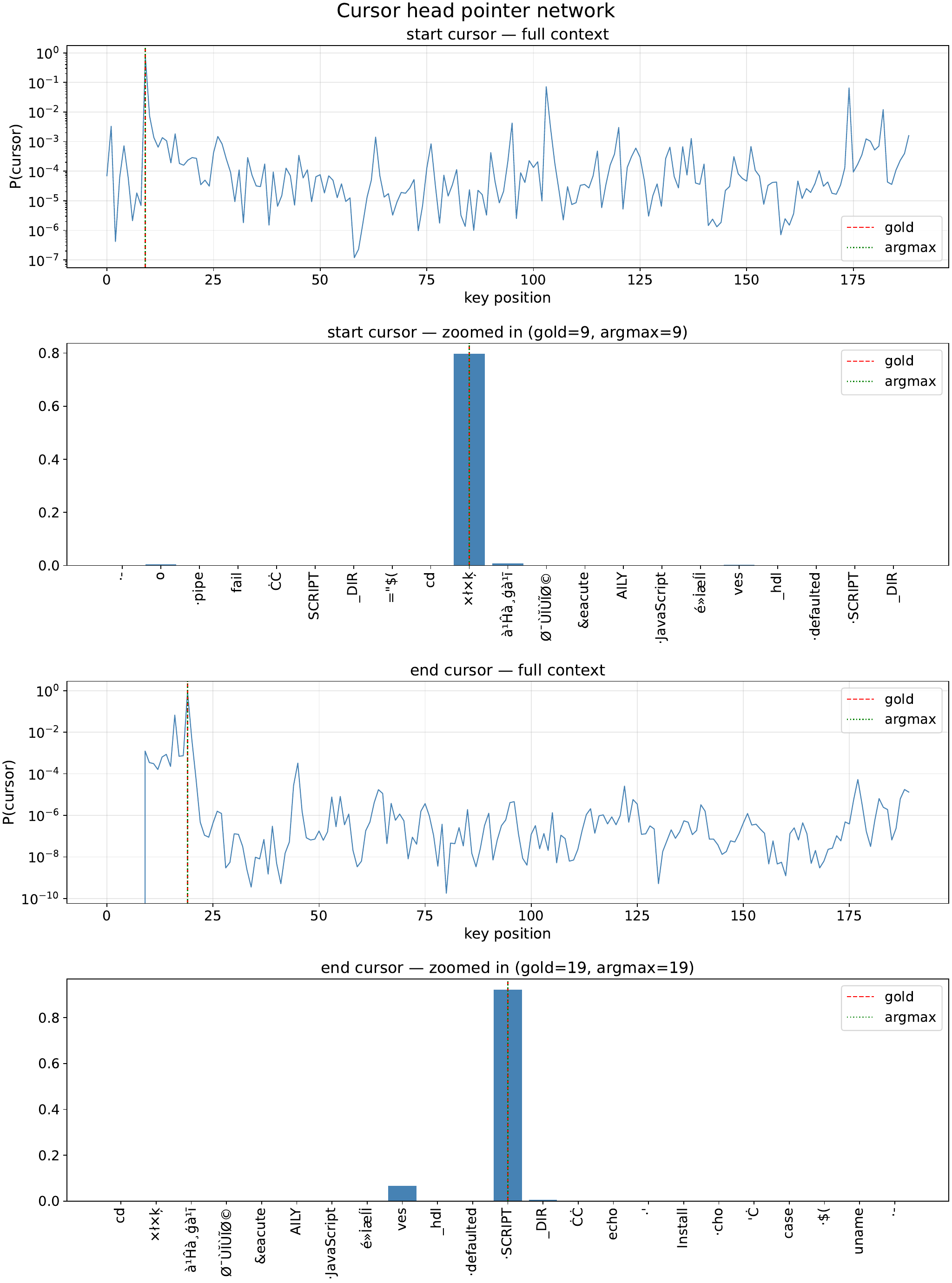}
    \caption{Visualization of the cursor pointer network heads for edit start and edit end. The edit start head attends to all tokens in the previous context to identify the beginning of a potential erroneous content. The edit end head attends to all tokens following the edit start cursor to identify where the edit should end.
    } 
     \label{fig:cursor_pointer_network}
\end{figure}

\end{document}

%% file: math_commands.tex
\usepackage{amsmath,amsfonts,bm}

\def\eqref#1{equation~\ref{#1}}

\def\1{\bm{1}}

\def\vtheta{{\bm{\theta}}}

\def\vk{{\bm{k}}}

\def\vq{{\bm{q}}}

\def\vx{{\bm{x}}}

\DeclareMathAlphabet{\mathsfit}{\encodingdefault}{\sfdefault}{m}{sl}
\SetMathAlphabet{\mathsfit}{bold}{\encodingdefault}{\sfdefault}{bx}{n}

\newcommand{\Ls}{\mathcal{L}}



%% file: tables/benchmarks_pre-training.tex
\begin{table}[t]
\centering
\caption{Pre-training benchmark results. All models were evaluated under the same protocol.
Numbers in parentheses are the number of few-shot examples used for in-context learning. 
The models with $^{\ast}$ were finetuned with 100K additional CPT training samples ($\sim$200M tokens). \textbf{Bold} shows the best score out of flexible-order decoding models, and the \underline{underline} shows the best overall.
}
\label{tab:benchmark-results}
\resizebox{\textwidth}{!}{%
\begin{tabular}{lcc|ccc}
\toprule
\textbf{Model} & \cellcolor{gray!10}\textbf{Qwen3 8B} & \cellcolor{gray!10}\textbf{Qwen3 8B LoRA}$^{\ast}$ & \textbf{LLaDA 8B} & \textbf{Dream 7B} & \textbf{AutoIndexer 8B LoRA}$^{\ast}$ \\
\midrule
Decoding strategy & \cellcolor{gray!10}AR & \cellcolor{gray!10}AR & Diffusion & Diffusion & CoE (Chain of Edits) \\
Editable context & \cellcolor{gray!10}No & \cellcolor{gray!10}No & Limited & Limited & Yes \\
\midrule
\multicolumn{6}{c}{General Tasks} \\
\midrule
MMLU (5) & \cellcolor{gray!10}76.7 & \cellcolor{gray!10}\underline{77.0} & 65.9 & 69.5 & \textbf{76.9} \\
ARC-Challenge (0) & \cellcolor{gray!10}56.7 & \cellcolor{gray!10}\underline{60.8} & 45.3 & \textbf{59.7} & 59.6 \\
Hellaswag (0) & \cellcolor{gray!10}\underline{78.7} & \cellcolor{gray!10}77.7 & 69.4 & 72.5 & \textbf{77.5} \\
WinoGrande (5) & \cellcolor{gray!10}\underline{76.9} & \cellcolor{gray!10}76.2 & 70.4 & 72.7 & \textbf{76.1} \\
PIQA (0) & \cellcolor{gray!10}79.6 & \cellcolor{gray!10}\underline{80.4} & 70.4 & 72.0 & \textbf{79.8} \\
\midrule
\multicolumn{6}{c}{Mathematics \& Science} \\
\midrule
GSM8K (5) & \cellcolor{gray!10}\underline{85.5} & \cellcolor{gray!10}85.0 & 71.3 & 75.1 & \textbf{84.3} \\
MATH500 (4) & \cellcolor{gray!10}55.2 & \cellcolor{gray!10}\underline{56.4} & 31.4 & 42.8 & \textbf{55.0} \\
\midrule
\multicolumn{6}{c}{Code} \\
\midrule
HumanEval (0) & \cellcolor{gray!10}45.7 & \cellcolor{gray!10}51.2 & 10.4 & 0.6 & \underline{\textbf{59.2}} \\
\quad -- with instruction & \cellcolor{gray!10}\underline{81.7} & \cellcolor{gray!10}81.1 & 36.0 & 56.7 & \textbf{79.3} \\
MBPP (3) & \cellcolor{gray!10}\underline{68.4} & \cellcolor{gray!10}67.6 & 39.4 & 54.4 & \textbf{65.8} \\
\bottomrule
\end{tabular}%
}
\vspace{-0.2em}
\end{table}

%% file: tables/benchmarks_post-training.tex
\begin{table}[t]
\centering
\caption{Post-training (instruction-tuned) benchmark results. All models were evaluated
instruct-formatted, under the same protocol.
Numbers in parentheses are the number of few-shot examples used for in-context learning.
The models with $^{\ast}$ were supervised-finetuned with the same 100K. \textbf{Bold} shows the best score out of flexible-order decoding models, and the \underline{underline} shows the best overall.
}
\label{tab:benchmark-results-sft}
\resizebox{\textwidth}{!}{%
\begin{tabular}{lcc|ccc}
\toprule
\textbf{Model} & \cellcolor{gray!10}\textbf{Qwen3 8B} & \cellcolor{gray!10}\textbf{Qwen3 8B LoRA}$^{\ast}$ & \textbf{LLaDA 8B Instruct} & \textbf{Dream 7B Instruct} & \textbf{AutoIndexer 8B LoRA}$^{\ast}$ \\
\midrule
Decoding strategy & \cellcolor{gray!10}AR & \cellcolor{gray!10}AR & Diffusion & Diffusion & CoE (Chain of Edits) \\
Editable context & \cellcolor{gray!10}No & \cellcolor{gray!10}No & Limited & Limited & Yes \\
\midrule
\multicolumn{6}{c}{General Tasks} \\
\midrule
ARC-Challenge (0) & \cellcolor{gray!10}90.61 & \cellcolor{gray!10}\underline{90.70} & 86.26 & 88.23 & \textbf{90.61} \\
MMLU-Pro (0) & \cellcolor{gray!10}42.38 & \cellcolor{gray!10}\underline{47.90} & 32.65 & 44.39 & \underline{\textbf{47.90}} \\
\midrule
\multicolumn{6}{c}{Mathematics \& Science} \\
\midrule
GPQA-Diamond (5) & \cellcolor{gray!10}35.35 & \cellcolor{gray!10}30.81 & 20.71 & \underline{\textbf{35.86}} & 30.81 \\
GSM8K (5) & \cellcolor{gray!10}\underline{86.96} & \cellcolor{gray!10}82.11 & 75.74 & 80.59 & \textbf{80.67} \\
MATH500 (4) & \cellcolor{gray!10}\underline{52.20} & \cellcolor{gray!10}49.20 & 17.00 & 45.80 & \textbf{49.00} \\
\midrule
\multicolumn{6}{c}{Code} \\
\midrule
HumanEval (0) & \cellcolor{gray!10}\underline{83.54} & \cellcolor{gray!10}73.17 & 45.12 & 52.44 & \textbf{75.61} \\
MBPP (3) & \cellcolor{gray!10}\underline{67.40} & \cellcolor{gray!10}7.60 & 37.60 & 58.00 & \textbf{64.00} \\
\bottomrule
\end{tabular}%
}
\vspace{-1em}
\end{table}

%% file: tables/teacher-forced-edits.tex
\begin{table}[!b]
    \caption{Teacher-forced edit execution with oracle-placed markers (window of 512 tokens). Measures the increase in $P(\textsc{edit})$ at the synthetic trigger position against the same position in a clean sequence, as well as the edit start and end localization, content improvement, and edit termination accuracy.}
    \label{tab:edit-teacher-forced}
    \centering
    \small
    \resizebox{\textwidth}{!}{%
    \begin{tabular}{lccc}
        \toprule
        Measurement & Insert & Delete & Substitute \\
        \midrule
        $P(\textsc{edit})$ at trigger (corrupted / clean) & 0.045 / 0.027 & 0.088 / 0.026 & 0.090 / 0.027 \\
        \quad -- Elevation over clean control & $\sim$1.7$\times$ & $\sim$3.4$\times$ & $\sim$3.4$\times$ \\
        \quad -- Paired win-rate ($P(\textsc{edit}|\textsc{corrupt}) > P(\textsc{edit}|\textsc{clean})$) & 90\% & 100\% & 100\% \\
        Correct edit start & 40\% & 76\% & 84\% \\
        Correct span (both start and end) & 38\% & 69\% & 74\% \\
        Replacement token accuracy & 76\% & -- & 80\% \\
        $P(\textsc{return})$ at correct stop & 72\% & 84\% & 73\% \\
        \bottomrule
    \end{tabular}%
    }
    \vspace{-1em}
\end{table}